# Cross-spectral Iris Recognition for Mobile Applications using High-quality Color Images

Mateusz Trokielewicz[1,2] and Ewelina Bartuzi[1]

[1] Biometrics Laboratory, Research and Academic Computer Network, Warsaw, Poland
[2] Institute of Control and Computation Engineering, Warsaw University of Technology, Warsaw, Poland

**Abstract**—With the recent shift towards mobile computing, new challenges for biometric authentication appear on the horizon. This paper provides a comprehensive study of cross-spectral iris recognition in a scenario, in which high quality color images obtained with a mobile phone are used against enrollment images collected in typical, near-infrared setups. Grayscale conversion of the color images that employs selective RGB channel choice depending on the iris coloration is shown to improve the recognition accuracy for some combinations of eye colors and matching software, when compared to using the red channel only, with equal error rates driven down to as low as 2%. The authors are not aware of any other paper focusing on cross-spectral iris recognition is a scenario with near-infrared enrollment using a professional iris recognition setup and then a mobile-based verification employing color images.

## 1. Introduction

### 1.1. Iris Biometrics

In the recent decades, biometric authentication and identification of humans has received considerable interest as a fast, safe, and convenient way of replacing password, token, or key-based security measures. One of the most accurate biometric methods is iris recognition, whose concept was first conceived by British ophthalmologists Safir and Flom [1] and later patented and implemented by Daugman [2], [3]. The iris, a part of the uvea, is located at the back of the anterior chamber of the eye, protected from the outside environment by eyelashes, tear film, the cornea and the aqueous humor. Its usefulness as a biometric identifier is attributed to the intricate patterns of the trabecular meshwork found in the front part of the organ. These patterns are developed in the embryonic stage and have low genotype dependence, thus providing enough unique features for high confidence classification. Iris texture is also believed to be exceptionally stable in time and virtually impossible to alter without inflicting extensive damage to the eye.

### 1.2. Cross-Spectral Iris Recognition

Iris recognition biometric systems are usually taking advantage of images collected using near-infrared (NIR) illumination. This is due to certain light absorption properties of melanin – the pigment, to which the iris attributes its appearance. While the absorption is significant for light from the visible spectrum, it is almost negligible for higher wavelengths. Higher reflectance enables good visibility of iris texture details even for highly pigmented (dark) irises. For this reason most commercial iris cameras collect images under illumination from the 700–900 nm wavelength range. However, with the recent shift in consumer computing towards mobile devices, visible light and cross-spectral iris recognition has received considerable attention.

This study aims at analyzing cross-spectral performance of the state-of-the-art iris recognition methods when applied with visible spectrum and near-infrared images. To our best knowledge, this is the first analysis of such kind incorporating high quality, flash-illuminated visible light iris images obtained using a mobile phone, which are then matched against NIR-illuminated enrollment samples. If good performance of the recognition methods can be achieved, it could pave the way for low-effort, real world applications, such as user authentication on a mobile device, which would serve as a remote verification terminal complementing a typical enrollment setup employing a NIR camera. We envisage a scenario, in which the enrollment stage is performed using a professional NIR setup for purposes such as government-issued IDs, travel documents, etc. Then, mobile authentication could be performed using a phone or a tablet whenever user deems it necessary. As of August 2016, there are only three iris recognition enabled mobile phones with the capability to obtain iris images in near infrared: Fujitsu NX F-04G ([4], available only in Japan), Microsoft Lumia 950 and 950XL ([5], employing Windows Hello iris-based authentication, currently in beta) and Samsung Galaxy Note 7 [6]. Due to this limited availability of NIR iris scanning equipment in mobile devices, a possibility of cross-spectral iris matching using color images is explored. Therefore, this paper offers three main contributions:

- evaluation of cross-spectral iris comparisons in a mobile recognition scenario,

- analysis of how the eye color influences cross-spectral iris matching,
- selection of the most efficient RGB channel depending on the eye color in order to optimize cross-spectral iris matching.

This paper is organized as follows: Section 2 presents an overview of the current state-of-the-art research related to this topic. Section 3 offers a brief explanation how the iris color in the human eye is determined. Multispectral database of iris images, data subset creation in respect to the eye color, and image preprocessing using selective RGB channel grayscale transformation are described in Section 4. Experimental methodology and software tools are characterized in Section 5, while Section 6 presents an overview of results. Finally, relevant conclusions are drawn in Section 7.

## 2. Related work

Probably the first systematic study devoted to multispectral iris biometrics was this of Boyce *et al.* [7], in which authors studied the reflectance response of the iris tissue depending on the spectral channel employed: red, green, blue, and infrared. Results of matching performance evaluation across channels and wavelengths are reported with the conclusion that decreases in matching accuracy is smallest in spectral channels that are closest to each other wavelengthwise. Technique for improving the recognition accuracy by employing histogram equalization in the CIE L*a*b color space is shown. Authors also present insight on how iris recognition could benefit from multispectral fusion in both segmentation and matching domains. Park *et al.* [8] explored fusing multispectral iris information as a countermeasure against spoofing. Iris features extracted from images acquired almost simultaneously both in low wavelength band and in high wavelength band are fused together in an attempt to create a method that would differentiate between real and counterfeit samples without compromising the recognition accuracy. Spectral variations found in real images obtained under different illumination conditions are said to offer enough variability to achieve this.

Ross *et al.* [9] were the first to investigate multispectral iris recognition using wavelengths longer than 900 nm, proving that images obtained in the range of 900–1400 nm can offer iris texture visibility good enough for biometric applications. At the same time, authors argue that the iris is able to give different responses as different wavelengths, which could prove useful for improving segmentation algorithms. Intra- (i.e. between images obtained in the same wavelength) and inter-spectral (i.e. between images obtained in two different wavelengths) genuine and impostor comparisons were generated. This revealed that despite inter-spectral genuine comparison distributions being shifted towards relevant impostor distributions, nearly perfect separation between genuine and impostor distributions can be achieved with the use of multispectral fusion at the score level.

Burge *et al.* [10] studied the iris texture appearance depending on the eye color combined with light wavelength employed for imaging. A method of approximating NIR image from visible light image is presented, together with multispectral iris fusion designed to create a high confidence image that would improve cross-spectral matching accuracy. Zuo *et al.* [11] attempted to predict NIR images from color images using predictive image mapping, to compare them against the enrolled typical NIR image. This method is said to outperform matching between NIR channel and red channel by roughly 10%.

Recently, advancements have also been made in the field of visible light iris recognition applications in more practical scenarios, including mobile devices – smartphones and tablets. Several databases have been released, including the UPOL database of iris images obtained using an ophthalmology device [12], the UBIRISv1 database, representing images obtained using Nikon E5700 camera and the UBIRISv2 database, which gathers images captured *on-the-move* and *at-a-distance* [13], [14]. These databases represent images obtained in very unconstrained conditions, and therefore usually of low quality. Recently, we have published the first available to researchers database of high quality iris images: the Warsaw-BioBase-Smartphone-Iris dataset [15], which comprises images obtained with iPhone 5s phone with embedded flash illumination (available online at [16]).

Challenges related to visible light iris recognition were extensively studied by Proenca *et al.*, including the amount of information that can be extracted from such images [17], possible improvements to the segmentation stage [18], and methods for image quality assessment to discard samples of exceptionally poor quality [19]. Santos *et al.* explored possible visible light illumination setups in the search for optimal solution for unconstrained iris acquisition [20]. Segmentation of noisy, low quality iris images was also studied by Radu *et al.* [21] and Frucci *et al.* [22]. Raja *et al.* explored visible spectrum iris recognition using a light field camera [23] and white LED illumination [24]. They also investigated a possibility of deploying iris recognition onto mobile devices using deep sparse filtering [25] and K-means clustering [26], reporting promising results such as EER as low as 0.31% when visible spectrum, smartphone-obtained images are used for recognition. The feasibility of face and iris biometrics implementations in mobile devices was also studied by De Marsico *et al.* [27]. Our own experimentations devoted to this field of research have shown that intra-wavelength visible spectrum iris recognition is possible when high-quality, color iris images obtained using a modern smartphone are used with the existing state-of-the-art methods, which are typically designed for NIR images [15], [28].

## 3. Anatomical Background of Iris Color

The iris consists of two major layers: the outer stroma, a meshwork of interlacing blood vessels, collagen fibers,

and sometimes melanin particles, and the inner epithelium, which connects to the muscles that control the pupil aperture. The epithelium itself contains dark brown pigments regardless of the observed eye color. The eye color perceived by the human observer depends mainly on the amount of melanin that can be found in the stroma. The more melanin in the stroma, the darker the iris appears due to absorption of incoming light by melanin. Blue hue of the iris, however, is attributed to light being scattered by the stroma, with more scattering occurring at higher frequencies, hence the color blue. This phenomenon, called the Tyndall effect, is similar to Rayleigh scattering and occurs in colloidal solutions, where the scattering particles are smaller than the scattered light wavelengths. Green eye color, on the other hand, is a result of combining these two phenomena (melanin light absorption and Tyndall scattering) [29].

## 4. Multispectral Database of Iris Images

For the purpose of this study a multispectral database of iris images has been collected, comprising NIR-illuminated images of standard quality (as recommended by ISO/IEC standard regarding biometric sample quality [30]) and high quality images obtained in visible light. 36 people presenting 72 different irises participated in the experiment. IrisGuard AD100, a two-eye NIR iris recognition camera, has been employed to capture six NIR-illuminated images (480 × 640 pixel bitmaps). Color images were acquired using the rear camera of Apple iPhone 5s (8-megapixel, JPG-compressed), with flash enabled. Data acquisition with a phone produced at least three images for each eye. In total, 432 near-infrared images acquired by a professional iris recognition camera and 272 color photos taken with a mobile phone were collected.

Images were then divided into three separable groups in respect to the eye color: the blue eyes subset, the green eyes subset, and the brown/hazel eyes subset, comprising 32 blue eyes, 18 green eyes and 22 brown, hazel or mixed-color eyes, respectively. Commercially available iris recognition software is typically built to cooperate with data compatible with the ISO/IEC standard specification. Color images were thus cropped to VGA resolution (640 × 480 pixels) and then converted to grayscale using selective RGB channel decomposition. Red, green, and blue channel of the RGB color space were extracted separately for each of the three eye color groups. Figure 1 presents sample images obtained using both cameras employed in this study, together with images extracted from each of the three RGB channels.

## 5. Experimental Methodology

### 5.1. Iris Recognition Software

For the most comprehensive analysis, three commercial, state-of-the-art iris recognition methods and one algorithm of academic origin have been employed. This section briefly characterizes each of these solutions.

Monro Iris Recognition Library (**MIRLIN**) is a commercially available product, offered on the market by FotoNation (formerly SmartSensors) [31] as an SDK (Software Development Kit). Its methodology incorporates calculating binary iris code based on the output of a discrete cosine transform (DCT) applied to overlapping iris image patches [32]. The resulting binary iris templates are compared using XOR operation and comparison scores are generated in the form of fractional Hamming distance, i.e. the proportion of disagreeing bits in the two iris codes. With this metric in place, we should expect values close to zero for genuine (i.e. same-eye) comparisons, and values around 0.5 for impostor (i.e. different-eye) comparisons. The latter is due to the fact that comparing bits in iris codes of two different irises can be depicted as comparing two sequences of independent Bernoulli trials (such as symmetric coin tosses).

**IriCore** employs a proprietary and unpublished recognition methodology. Similarly to MIRLIN, it is offered on the market in the form of an SDK by IriTech [33]. With this matcher, values between 0 and 1.1 should be expected for same-eye comparisons, while different-eye comparisons should yield scores between 1.1 and 2.0.

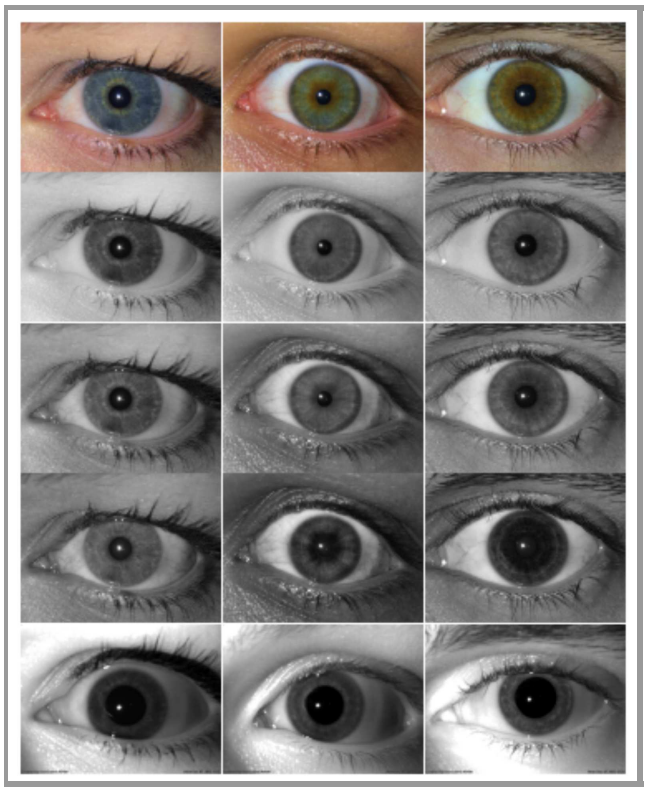

**Fig. 1.** Top row: cropped sample images obtained with the iPhone 5s: blue eye (left), green eye (middle), brown/hazel eye (right). Rows 2–4: grayscale versions of the images using red, green and blue channel of the RGB color space, respectively. Bottom row: NIR images obtained with the IrisGuard AD100. (See color pictures online at www.nit.eu/publications/journal-jtit)

The third method involved in this study, **VeriEye**, is available commercially from Neurotechnology [34] and, similarly to the IriCore method, the precise mechanisms of the recognition methodology are not disclosed in any scientific papers. The manufacturer, however, claims to employ active shape modeling for iris localization using non-circular approximations of pupillary and limbic iris boundaries. VeriEye, contrary to two previous methods, returns comparison scores in a form of similarity metric – the higher the score, the better the match. A perfect non-match should return a score equal to zero.

The last method employed for the purpose of this study is Open Source for IRIS (**OSIRIS**), developed within the BioSecure project [35] and offered by its authors as a free, open-source solution. OSIRIS follows the well-known concept originating in the works of Daugman, incorporating image segmentation and normalization by unwrapping the iris image from polar coordinates onto a Cartesian rectangle using Daugman's *rubber sheet* model. Encoding of the iris is carried out using phase quantization of multiple Gabor wavelet filtering outcomes, while matching is performed using XOR operation, with normalized Hamming distance as an output dissimilarity metric. As in the MIR-LIN method, values close to zero are expected for genuine comparisons, while impostor comparisons should typically produce results around 0.5, however, due to shifting the iris code in search for the best match as a countermeasure against eye rotation, impostor score distributions will more likely be centered around 0.4 to 0.45 values.

### 5.2. Comparison Scores Generation

As this study aims at quantifying cross-spectral iris recognition accuracy in a scenario that would mimic potential real-world applications, where mobile-based verification would complement a typical enrollment using professional iris recognition hardware operating in NIR, the following experiments are performed. NIR images obtained using the IrisGuard AD100 camera are used as gallery (enrollment) samples. Visible light images obtained with the iPhone 5s are used as probe (verification) samples. All possible genuine and impostor comparisons are generated for all three subsets of eyes and all three RGB channels, for each of the four iris recognition methods employed. Thus, 36 Receiver Operating Characteristic (ROC) curves can be constructed (4 methods × 3 eye color subsets × 3 RGB channels), 9 for each method. These are presented and commented on in the following section.

## 6. Results

Figures 2–5 illustrate ROC curves obtained when generating genuine and impostor score distributions for each RGB channel and each eye color subset. EER-wise, the red channel offers the best performance in most of the method/channel/subset combinations. There are however, a few exceptions from this behavior. For the blue eyes subset, the green channel provides recognition accuracy that is very similar to this of the red channel (slightly better for the MIRLIN matcher, slightly worse for the VeriEye matcher, and the same for the remaining two methods). Interestingly, for the MIRLIN matcher, the blue channel gives the same EER as the green channel, and better than the red channel. For the green eyes subset, the red channel offers significantly better performance than the other channels for the OSIRIS and MIRLIN matchers. However, for the VeriEye matcher, using green channel instead decreased the EER from 5 to 2%, a significant improvement over the recognition accuracy offered by the red channel.

The brown/hazel eyes subset, unsurprisingly, achieves the optimal recognition performance for the red channel, as it offers significantly better iris pattern visibility than the other two channels. The IriCore method, however, seems to be less susceptible to the type of the input data, as decrease in performance for the green channel is much lower than in the remaining three methods (compared to the red channel).

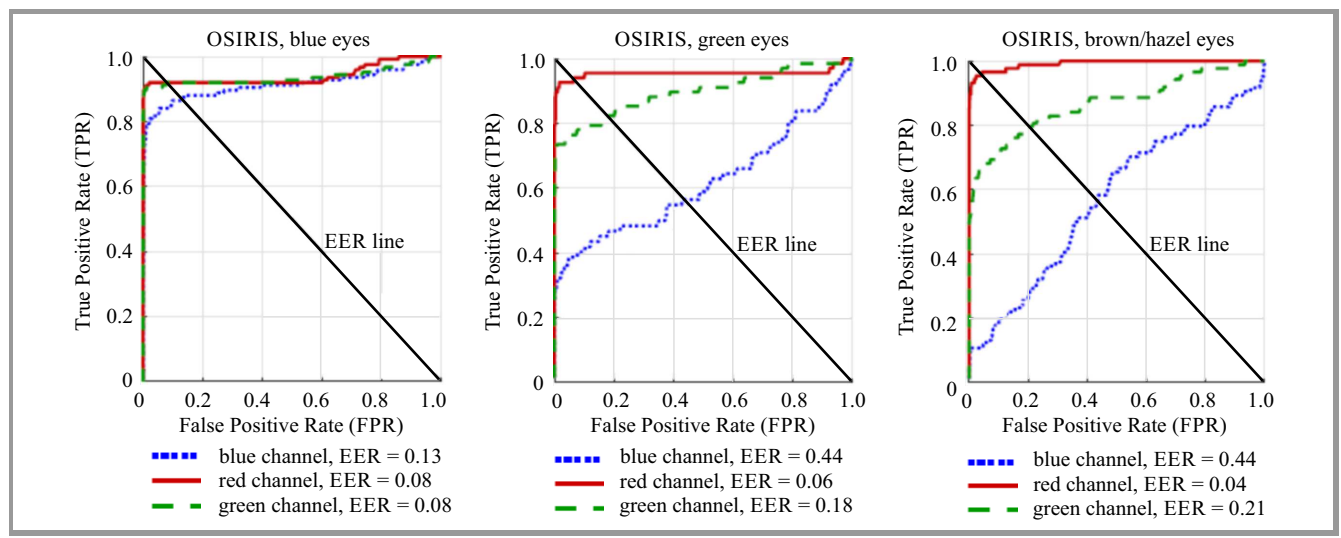

**Fig. 2.** ROC curves for **OSIRIS** matcher and scores obtained when matching different RGB channels of samples from the: (a) blue eyes, (b) green eyes, and (c) brown/hazel eyes subsets. Red channel scores are denoted with solid red line, blue channel scores: dotted blue line, green channel scores: dashed green line. Equal error rates (EER) are also shown.

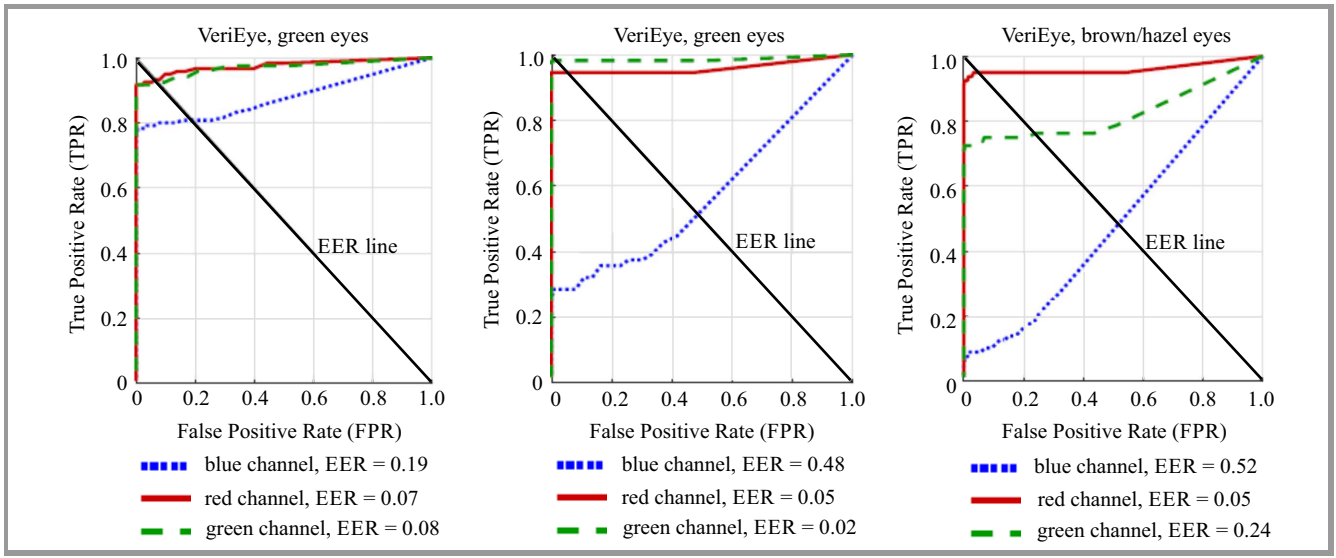

***Fig. 3.*** Same as in Fig. 2, but for the VeriEye matcher: (a) blue eyes, (b) green eyes, and (c) brown/hazel eyes.

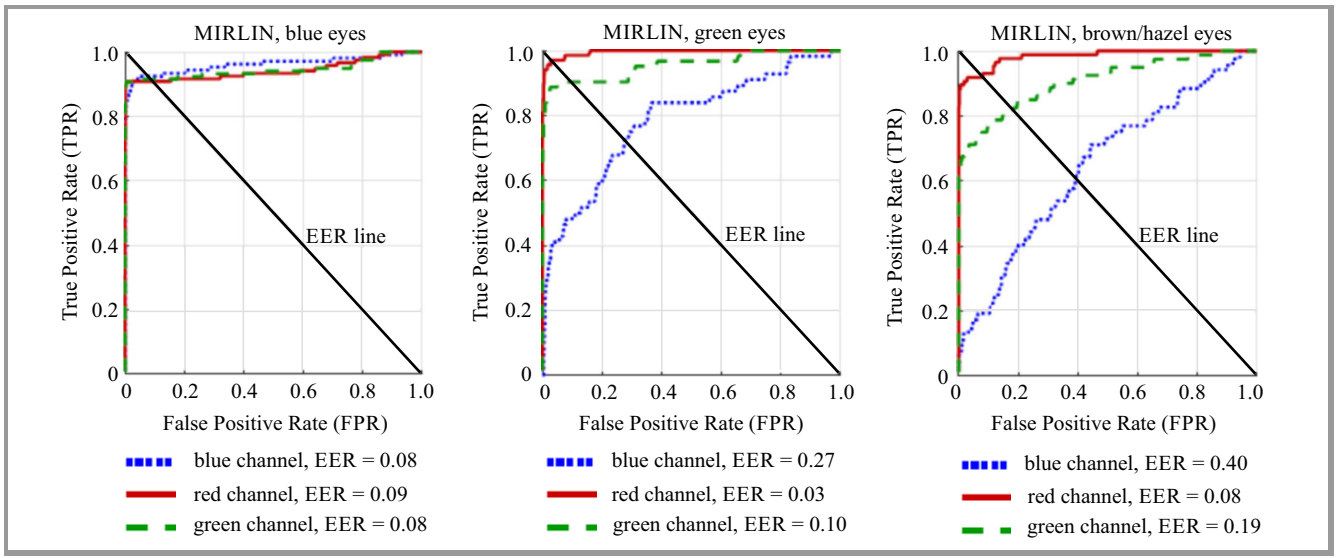

***Fig. 4.*** Same as in Fig. 2, but for the MIRLIN matcher: (a) blue eyes, (b) green eyes, and (c) brown/hazel eyes.

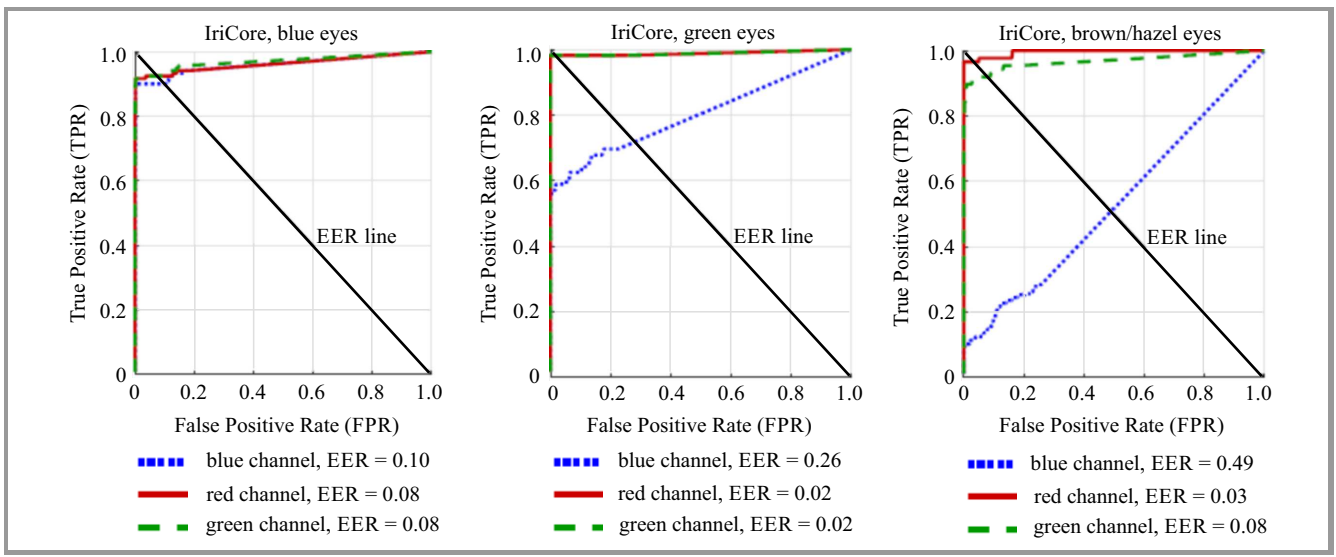

***Fig. 5.*** Same as in Fig. 2, but for the IriCore matcher: (a) blue eyes, (b) green eyes, and (c) brown/hazel eyes.

# 7. Conclusions

This study provides a valuable analysis of cross-spectral iris recognition, when high quality visible light images obtained with a mobile phone are used as verification counterparts for enrollment samples obtained in NIR. The red channel performs best in general, however, in selected cases, employing the green or the blue channel of the RGB color space when converting the color image to grayscale is shown to improve recognition accuracy. This is true for some combination of recognition methods and eye colors recognition accuracy can be improved this way in blue and green eyes, while dark brown and hazel eyes generally perform best when the red channel is used.

The experiments revealed that cross-spectral iris recognition in the discussed scenario is perfectly viable, with equal error rates not exceeding 7, 2 and 3% for the blue, green and brown/hazel eyes, respectively, when optimal combinations of the recognition method and grayscale transformation are selected. As this incorporated only a simple selection of the RGB channel best suited for a given eye color, future experiments employing more advanced image processing could bring the error rates even lower. This certainly lets us think of cross-spectral iris recognition using mobile phones with a great dose of optimism.

# Acknowledgement

The authors would like to thank Dr. Adam Czajka for his valuable comments that contributed to the quality of this paper. We are also grateful for the help of the Biometrics Scientific Club members at the Warsaw University of Technology when building the database used in this study.

# References

[1] L. Flom and A. Safir, "Iris recognition system", United States Patent, US 4641349, 1987.

[2] J. Daugman, "Biometric personal identification system based on iris analysis", United States Patent, US 5291560, 1994.

[3] J. G. Daugman, "High confidence visual recognition of persons by a test of statistical independence", *IEEE Trans. Pattern Anal. and Machine Intell.*, vol. 15, no. 11, pp. 1148–1161, 1993.

[4] Fujitsu Limited, Fujitsu Develops Prototype Smartphone with Iris Authentication [Online]. Available: http://www.fujitsu.com/global/about/resources/news/press-releases/2015/0302-03.html (accessed on Oct. 29, 2015).

[5] Planet Biometrics, Microsoft brings iris recognition to the masses with new Lumia [Online]. Available: http://www.planetbiometrics.com/article-details/i/3606/desc/microsoft-brings-iris-recognition-to-the-masses-with-new-lumia/ (accessed July 31, 2016).

[6] Samsung Electronics, Samsung Galaxy Note 7 [Online]. Available: www.samsung.com/global/galaxy/galaxy-note-7/security (accessed Aug. 10, 2016).

[7] C. Boyce, A. Ross, M. Monaco, L. Hornak, and X. Li, "Multispectral iris analysis: A preliminary study", in *Proc. Conf. Comp. Vision & Pattern Recogn. Worksh. CVPRW'06*, New York, NY, USA, 2006 (doi: 10.1109/CUPRW.2006.141).

[8] J. H. Park and M. G. Kang, "Multispectral iris authentication system against counterfeit attack using gradient-based image fusion", *Optical Engin.*, vol. 46, no. 11, 2007 (doi: 10.1117/1.2802367).

[9] A. Ross, R. Pasula, and L. Hornak, "Iris recognition: On the segmentation of degraded images acquired in the visible wavelength", in *Proc. IEEE 3rd Int. Conf. Biometrics: Theory, Appl. and Syst. BTAS 2009*, Washington, DC, USA, 2009.

[10] M. J. Burge and M. K. Monaco, "Multispectral iris fusion for enhancement, interoperability, and cross wavelength matching", in *Algorithms and Technologies for Multispectral, Hyperspectral, and Ultraspectral Imagery XV*, S. S. Shen and P. E. Lewis, Eds. *Proc. of SPIE*, vol. 7334, 73341D, 2009 (doi: 10.1117/12.819058).

[11] J. Zuo, F. Nicolo, and N. A. Schmid, "Cross spectral iris matching based on predictive image mapping", in *4th IEEE Int. Conf. Biometrics: Theory, Appl. and Syst. BTAS 2010*, Washington, DC, USA, 2010.

[12] M. Dobes, L. Machala, P. Tichavsky, and J. Pospisil, "Human eye iris recognition using the mutual information", *Optik*, vol. 115, no. 9, pp. 399–404, 2004.

[13] H. Proença and L. A. Alexandre, "UBIRIS: A noisy iris image database", Tech. Rep., ISBN: 972-99548-0-1, University of Beira Interior, Portugal, 2005.

[14] H. Proença, S. Filipe, R. Santos, J. Oliveira, and L. A. Alexandre, "The UBIRIS.v2: A database of visible wavelength iris images captured on-the-move and at-a-distance", *IEEE Trans. Pattern Anal. and Machine Intell.*, vol. 32, no. 8, pp. 1529–1535, 2010.

[15] M. Trokielewicz, "Iris recognition with a database of iris images obtained in visible light using smartphone camera", in *Proc. IEEE Int. Conf. on Ident., Secur. and Behavior Anal. ISBA 2016*, Sendai, Japan, 2016.

[16] Warsaw-BioBase-Smartphone-Iris-v1.0 [Online]. Available: http://zbum.ia.pw.edu.pl/en/node/46

[17] H. Proença, "On the feasibility of the visible wavelength, at-a-distance and on-the-move iris recognition", in *Proc. IEEE Symp. Series on Computat. Intell. in Biometr.: Theory, Algorithms, & Appl. SSCI 2009*, Nashville, TN, USA, 2009, vol. 1, pp. 9–15.

[18] H. Proença, "Iris recognition: On the segmentation of degraded images acquired in the visible wavelength", *IEEE Trans. on Pattern Anal. & Mach. Intellig.*, vol. 32, no. 81 pp. 1502–1516, 2010.

[19] H. Proença, "Quality assessment of degraded iris images acquired in the visible wavelength", *IEEE Trans. Inform. Forens. and Secur.*, vol. 6, no. 1, pp. 82–95, 2011.

[20] G. Santos, M. V. Bernardo, H. Proenca, and P. T. Fiadeiro, "Iris recognition: Preliminary assessment about the discriminating capacity of visible wavelength data", in *Proc. 6th IEEE Int. Worksh. Multim. Inform. Process. and Retrieval MIPR 2010*, Taichung, Taiwan China, 2010, pp. 324–329.

[21] P. Radu, K. Sirlantzis, G. Howells, S. Hoque, and F. Deravi, "A colour iris recognition system employing multiple classifier techniques", *Elec. Lett. Comp. Vision and Image Anal.*, vol. 12, no. 2, pp. 54–65, 2013.

[22] M. Frucci, C. Galdi, M. Nappi, D. Riccio, and G. Sanniti di Baja, "IDEM: Iris detection on mobile devices", in *Proc. 22nd Int. Conf. Pattern Recogn. ICPR 2014*, Stockholm, Sweden, 2014.

[23] K. Raja, R. Raghavendra, F. Cheikh, B. Yang, and C. Busch, "Robust iris recognition using light field camera", in *The 7th Colour and Visual Comput. Symp. CVCS 2013*, Gjøvik, Norway, 2013.

[24] K. Raja, R. Raghavendra, and C. Busch, "Iris imaging in visible spectrum using white LED", in *7th IEEE Int. Conf. on Biometr.: Theory, Appl. and Syst. BTAS 2015*, Arlington, VA, USA, 2015.

[25] K. Raja, R. Raghavendra, V. Vemuri, and C. Busch, "Smartphone based visible iris recognition using deep sparse filtering", *Pattern Recogn. Lett.*, vol. 57, pp. 33–42, 2014.

[26] K. B. Raja, R. Raghavendra, and C. Busch, "Smartphone based robust iris recognition in visible spectrum using clustered K-mean features", in *Proc. IEEE Worksh. Biometr. Measur. and Syst. for Secur. and Med. Appl. BioMS 2014*, Rome, Italy, 2014, pp. 15–21.

[27] M. De Marsico, C. Galdi, M. Nappi, and D. Riccio, "FIRME: Face and iris recognition engagement", *Image and Vis. Comput.*, vol. 32, no. 12, pp. 1161–1172, 2014.

[28] M. Trokielewicz, E. Bartuzi, K. Michowska, A. Andrzejewska, and M. Selegrat, "Exploring the feasibility of iris recognition for visible spectrum iris images obtained using smartphone camera", in *Photonics Applications in Astronomy, Communications, Industry, and High-Energy Physics Experiments 2015*, R. S. Romaniuk, Ed. *Proc. of SPIE*, vol. 9662, 2015 (doi: 10.1117/12.2205913).

[29] P. van Slembrouck, "Structural Eye Color is Amazing" [Online]. Available: http://medium.com/@ptvan/structural-eye-color-is-amazing-24f47723bf9a (accessed Aug. 8, 2016).

[30] ISO/IEC 19794-6:2011. Information technology – Biometric data interchange formats – Part 6: Iris image data, 2011.

[31] Smart Sensors Ltd., MIRLIN SDK, version 2.23, 2013.

[32] D. M. Monro, S. Rakshit, and D. Zhang, "DCT-based iris recognition", *IEEE Trans. Pattern Anal. and Machine Intell.*, vol. 29, no. 4, pp. 586–595, 2007.

[33] IriTech Inc., IriCore Software Develope's Manual, version 3.6, 2013 [Online]. Available: www.iritech.com/products/swoftware/iricore-eye-recognition-software

[34] Neurotechnology Company, VeriEye SDK, version 4.3 [Online]. Available: www.neurotechnology.com/verieye.html (accessed Aug. 11, 2015).

[35] G. Sutra, B. Dorizzi, S. Garcia-Salitcetti, and N. Othman, "A biometric reference system for iris. OSIRIS version 4.1 [Online]. Available: http://svnext.it-sudparis.eu/svnview2-eph/ref_syst/iris_osiris_v4.1 (accessed Oct. 1, 2014).

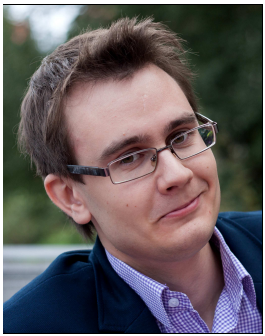

**Mateusz Trokielewicz** received his B.Sc. and M.Sc. in Biomedical Engineering from the Faculty of Mechatronics and the Faculty of Electronics and Information Technology at the Warsaw University of Technology, respectively. He is currently with the Biometrics Laboratory at the Research and Academic Computer Network and with the Institute of Control and Computation Engineering at the Warsaw University of Technology, where he is pursuing his Ph.D. in Biometrics. His current professional interests include iris biometrics and its reliability against biological processes, such as aging and diseases, and iris recognition on mobile devices.
E-mail: mateusz.trokielewicz@nask.pl
Biometrics Laboratory
Research and Academic Computer Network (NASK)
Kolska st 12
01-045 Warsaw, Poland

Institute of Control and Computation Engineering
Warsaw University of Technology
Nowowiejska st 15/19
00-665 Warsaw, Poland

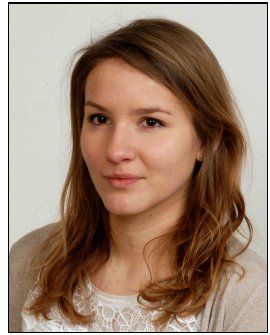

**Ewelina Bartuzi** received her B.Sc. degree in Biomedical Engineering from the Faculty of Mechatronics at the Warsaw University of Technology in 2016. She is presently working as a Technical Specialist at the Biometrics Laboratory, Research and Academic Computer Network. Her current research interests include biometrics based on thermal hand images and iris recognition for mobile.
E-mail: ewelina.bartuzi@nask.pl
Biometrics Laboratory
Research and Academic Computer Network (NASK)
Kolska st 12
01-045 Warsaw, Poland